\PassOptionsToPackage{noerr}{pdfx}
\documentclass[
]{ceurart}

\usepackage{listings}
\usepackage{booktabs}
\usepackage{multirow}
\usepackage{amsmath}
\usepackage{xspace}
\usepackage{float}
\usepackage[section]{placeins}
\usepackage{makecell}

\newcommand{\keep}{\textsc{Keep}\xspace}
\newcommand{\redact}{\textsc{Redact}\xspace}

\begin{document}

\copyrightyear{2026}
\copyrightclause{Copyright for this paper by its authors.
  Use permitted under Creative Commons License Attribution 4.0
  International (CC BY 4.0).}

\conference{AI for Education Day, SIGKDD 2026, August 12, 2026, Jeju, Korea}

\title{Redact or Keep? A Fully Local AI Cascade for Educational Dialogue De-Identification}

\author[1]{Haocheng Zhang}[%
  orcid=0009-0009-8580-1571,
  email=hz782@cornell.edu,
]
\author[1]{Zhuqian Zhou}[%
  orcid=0000-0002-8045-6213,
  email=zz968@cornell.edu,
]
\author[1]{Kirk Vanacore}[%
  orcid=0000-0003-0673-5721,
  email=kpv27@cornell.edu,
]
\author[1]{Bakhtawar Ahtisham}[%
  orcid=0009-0007-9331-5069,
  email=ba453@cornell.edu,
]
\author[1]{Ren\'e F. Kizilcec}[%
  orcid=0000-0001-6283-5546,
  email=kizilcec@cornell.edu,
]
\address[1]{Cornell University, Ithaca, NY, USA}

\begin{abstract}
Educational dialogue is a valuable but sensitive resource for research, yet de-identification is difficult because personally identifiable information often overlaps with curricular content: “Riemann” may refer to a student or to a mathematical concept. Commercial LLMs can handle this ambiguity but require data egress, while local NER systems preserve governance but often over-redact educational content. We propose a fully local cascade that separates de-identification into high-recall candidate proposal and contextual Redact/Keep review. The first stage combines two lightweight encoders with rule-based patterns to over-generate candidate spans; the second stage decides whether each span is real PII using surrounding dialogue and speaker role. Across math tutoring transcripts from two platforms, the strongest local configuration achieves a macro F1 of 0.958, compared with 0.767 for a same-family LLM-only baseline and 0.706 for a commercial API. On a targeted challenge set for curricular–personal name ambiguity, it degrades by only 0.03 F1 versus 0.19–0.25 for smaller reviewers. These results show that for educational de-identification, problem formulation can matter more than model scale, enabling privacy-preserving deployment without external APIs.
\end{abstract}

\begin{keywords}
  de-identification \sep
  PII detection \sep
  educational NLP \sep
  local deployment \sep
  name disambiguation
\end{keywords}

{\hfuzz=500pt\vfuzz=500pt\maketitle}

\section{Introduction}
\label{sec:intro}

Authentic educational dialogue is a rich but sensitive resource for research. The same transcripts that capture how students learn often capture what they disclose about themselves, requiring removal of personally identifiable information (PII) before release. In educational dialogue, de-identification is harder than standard named-entity recognition because the same surface form may refer either to curricular content that should be preserved (e.g., ``Riemann'' in ``Riemann sums'') or to a real student that must be redacted (``Hi, I'm Maria Riemann''). A practical system must therefore balance four requirements: detection accuracy, curricular preservation, privacy governance, and deployment cost.

Commercial LLM APIs improve detection performance \cite{singhal2024deid,ji2025enhancing,zhou2026mathedpii}, but sending student transcripts to third-party providers introduces data governance and compliance risks that many institutions worry about. Generic local named entity recognition (NER) systems avoid data egress but often fail on curricular-personal ambiguity. Existing approaches, therefore, trade governance for accuracy or vice versa.

To address these constraints, we propose a fully local cascade framework for educational dialogue de-identification. A \emph{union} proposer combines two lightweight encoders (DeBERTa and ModernBERT) with direct rules to intentionally over-generate candidate PII spans, optimizing for recall at the expense of precision (recall-first). A second-stage \emph{cascade-aligned reviewer} uses an LLM to inspect each candidate in context and emits a binary \redact/\keep decision. This formulation prioritizes recall-first but allows the reviewer to prevent over-reductions when necessary by converting open-ended entity recognition into constrained privacy triage. We compare three reviewer configurations---an encoder reviewer, a 4B LLM reviewer trained with Low-Rank Adaptation (LoRA), and a 31B LoRA-trained LLM reviewer---against same-family LLM-only baselines and a commercial API baseline.

This setting is a data-mining problem because the central challenge is not only detecting entities but also preserving analytic utility in large-scale educational dialogue while satisfying privacy constraints. The paper addresses the following research questions (RQs):
\begin{itemize}
\itemsep0em
\item \textbf{RQ1 — Cascade vs. LLM-Only.} Holding the base model family constant, does a cascaded proposer--reviewer framework outperform single-step LLM-only full-dialogue extraction?
\item \textbf{RQ2 — Robustness on ambiguous names.} On a pre-registered challenge set where real student names overlap with curricular-content-based names, which reviewers remain robust?
\item \textbf{RQ3 — Deployment footprint.} Under commodity hardware constraints, what are training and inference costs, and can the full pipeline remain fully on-device?
\end{itemize}

\section{Related Work}
\label{sec:prior-work}

\subsection{From Mature De-identification Domains to Educational Dialogue}

De-identification is well established in healthcare, where shared tasks such as i2b2/UTHealth have supported mature sequence-labeling approaches \cite{stubbs2015i2b2}. However, clinical assumptions do not transfer cleanly to educational dialogue: the same surface form may refer to a student, a public figure, or a fictional character in a word problem \cite{kovacevic2024systematic}. Educational de-identification is therefore less a pure span-detection task than a contextual privacy decision.

Educational de-identification often has substantially different affordances that require distinct approaches.  PIIvot \cite{zent2025piivot} addresses tutoring-dialogue anonymization through recall-first potential-PII labeling and surrogate replacement, but does not make the explicit \redact{}/\keep{} preservation decision required for curricular-content disambiguation. MathEd-PII \cite{zhou2026mathedpii} targets numeric ambiguity in mathematical expressions, rather than name disambiguation. \citet{ji2025enhancing} and \citet{singhal2024deid} evaluate GPT-4o-mini and GPT-4 on educational corpora \cite{holmes2024crapii}, achieving high accuracy but sending transcripts through commercial APIs, trading governance for accuracy. However, none of these evaluate the curricular-personal name ambiguity typical of educational contexts, nor do they report an end-to-end local deployment.

\subsection{From Entity Detection to Privacy Triage}

Off-the-shelf detectors like spaCy \cite{honnibal2020spacy} and Microsoft Presidio \cite{microsoft2024presidio} are local and fast but trained on news-like text. Their default behavior tends to over-redact curricular terms while providing no explicit keep-vs-redact distinction for ambiguous mentions. Customized Presidio recognizers and transformer-backed spaCy may improve lexical recall, but they still treat the task as entity detection rather than contextual privacy triage. In educational dialogue, identifying a candidate mention is often straightforward; deciding whether it should actually be redacted is the harder problem.

Two-stage NER systems provide a useful structural template, separating span proposal from subsequent classification or typing. Prior variants include cascaded fine-grained NER, T2-NER, TadNER, and ToNER \cite{awasthy2020cascaded, huang2023t2ner, li2023tadner, jiang2024toner}. In these systems, the second stage typically performs entity typing. Our work instead uses the second stage for privacy verification: deciding whether a candidate span should be redacted or preserved in context.

\section{Proposed Pipeline}
\label{sec:pipeline}

\begin{figure}[!ht]
\centering
\includegraphics[width=\linewidth]{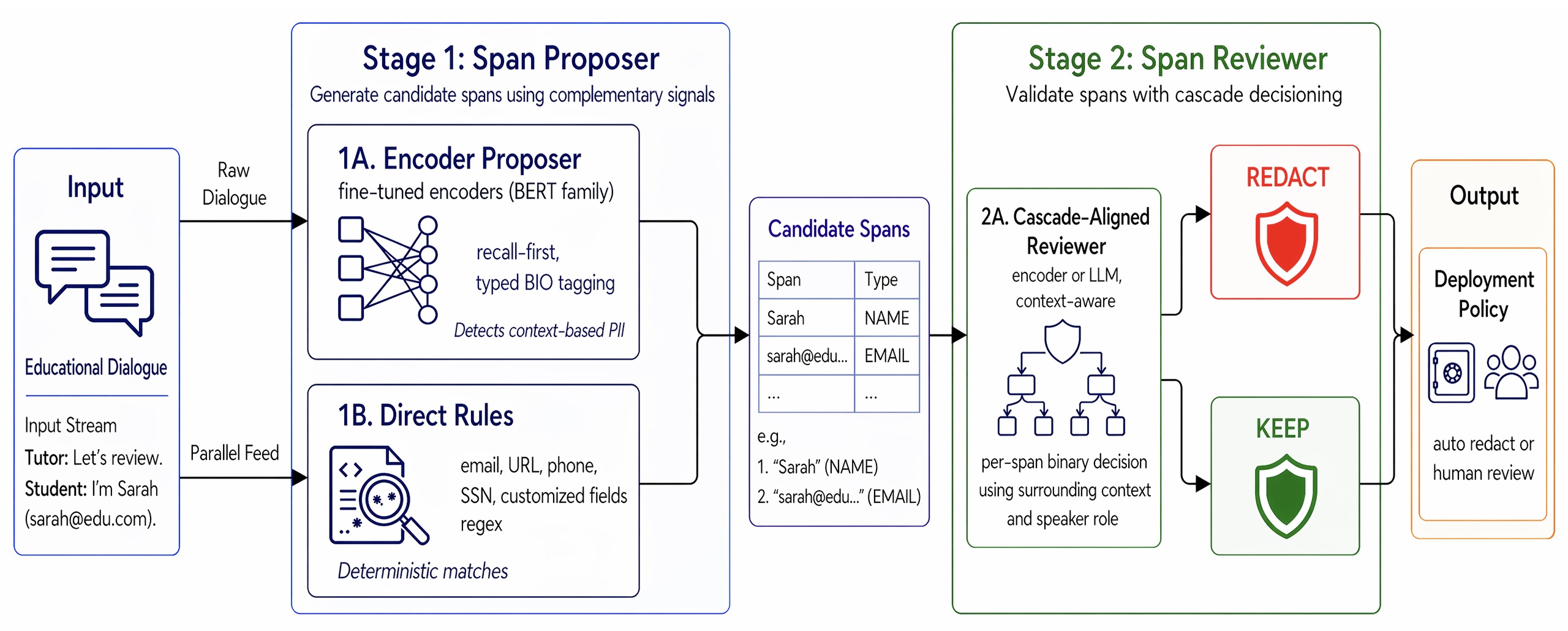}
\caption{End-to-end cascade pipeline. A recall-first \emph{union} encoder proposer (DeBERTa-v3-base + ModernBERT-base; stage 1A) plus direct rules (stage 1B) emit candidate spans; a cascade-aligned reviewer (stage 2A) decides \redact/\keep for each candidate using surrounding context and speaker role; a deployment-time policy applies the final privacy action.}
\label{fig:architecture}
\end{figure}

\subsection{Pipeline Overview}

As noted above, we formulate de-identification as a two-stage classification problem over dialogue turns. Figure~\ref{fig:architecture} illustrates the pipeline. Each dialogue turn is processed in parallel by the encoder proposer (a union of fine-tuned DeBERTa and ModernBERT; Stage~1A) and a set of direct RegEx rules (Stage~1B). The candidate pool is the union of spans from either source, intentionally favoring high recall over early precision. Each candidate span is then passed to the reviewer (Stage~2) together with its surrounding context, speaker role, and turn metadata, while rule-matched spans are high-precision structured identifiers and are emitted as \redact directly, bypassing the reviewer (Appendix~\ref{sec:methodology-notes}). Rather than performing open-ended entity extraction, the reviewer makes a constrained binary \redact or \keep decision. This reframes the task from entity recognition to privacy triage. Spans predicted as PII are labeled \redact and then handled according to a deployment-time policy. In an \emph{auto-redact} setting, all such spans are automatically redacted to minimize disclosure risk. In a \emph{human-review} setting, low-confidence cases are routed to manual verification, allowing institutions to balance privacy protection against curricular preservation. As constructed, the entire pipeline can be run locally with no external API calls.

\subsection{Stage 1: Span Proposer}
\label{sec:stage1}

The proposer is a recall-first union of two fine-tuned encoders, DeBERTa-v3-base \cite{he2023debertav3} and ModernBERT-base \cite{warner2024modernbert}, both trained with typed BIO token classification, a sequence-labeling schema that marks each token as the Beginning, Inside, or Outside of a typed entity span (e.g., B-\textsc{Name}, B-\textsc{Address}). Any span proposed by either encoder is passed forward, intentionally favoring recall over early precision.

In parallel, a deterministic rule-based proposer handles structured identifier types with strong lexical patterns, including \textsc{Email}, \textsc{URL}, \textsc{IP\_Address}, \textsc{Phone\_Number}, \textsc{Identifying\_Number}, and customized lexicon-based types (e.g., in our example, we also redacted the name of the \textsc{Tutor\_Provider}s that generated the data). These types are sparse in educational data and well-suited to high-precision pattern matching, making rules more reliable and efficient than learned detection.

\subsection{Stage 2: Cascade-Aligned Reviewer}

\subsubsection{Label Alignment}
\label{sec:label-alignment}

The reviewer operates on proposer-generated candidate spans rather than full dialogues. For each proposed span, overlap with a gold annotation (manually labeled PII spans; see Section~\ref{sec:data}) is mapped to \redact (true positive), while no overlap is mapped to \keep (false positive). This mechanically aligns training labels with the actual deployment decision: whether the span should be removed. This formulation converts open-ended NER into a constrained binary classification task. Instead of asking the model to extract all sensitive spans from raw dialogue, the reviewer only determines whether a proposed candidate is true PII. This reduces task complexity, improves alignment with privacy decisions, and makes small local models substantially more practical. Reviewer training uses binary labels with focal loss \cite{lin2017focal} to address class imbalance.

\subsubsection{Reviewer Configurations}

We evaluate three reviewer configurations on the same union-proposer output: a RoBERTa reviewer (125M, full fine-tune, 3 seeds) \cite{liu2019roberta}, a Gemma E4B-it \cite{gemma4_2026} reviewer (4B, LoRA  \cite{hu2022lora,dettmers2023qlora}, 1 seed), and a Gemma-4-31B-it \cite{gemma4_2026} reviewer (31B, LoRA, 3 seeds). Hyperparameters and the mlx-lm gradient-clipping patch required for 3-seed 31B training are in Appendix~\ref{sec:hyperparams}. Both the 4B and 31B LoRA LLM reviewers share a context-aware system prompt (Appendix~\ref{sec:reviewer-prompt}) that instructs the model to preserve names used in educational content while still redacting real student identities.

\section{Method}
\label{sec:experiments}

\subsection{Data}
\label{sec:data}

\subsubsection{Data Sources}

We evaluate the pipeline on math tutoring transcripts from two large online tutoring providers with different instructional settings. Table \ref{tab:dataset-summary} presents the platforms and their respective data sets. Platform A contains short, question-focused K--12 tutoring dialogues, while Platform B contains longer one-to-one K--12 tutoring dialogues from scheduled online lessons with extended multi-turn interaction. As a result, Platform B sessions are substantially longer and more conversational than Platform A. All transcripts were manually annotated for PII by three annotators using a shared codebook with disagreements resolved through consensus discussion.

\renewcommand{\cellalign}{cl}

\FloatBarrier
\begin{table}[!ht]
\caption{Dataset summary. ``Candidate turns'' are turns scored by the Stage-1 Span Proposer as potential turns with PII which are used as the input for the Stage-2 Span Reviewer. ``Dialogues without PII'' are retained as negative examples.}
\centering
\small
\setlength{\tabcolsep}{3pt}
\begin{tabular}{@{}lccccccc@{}}
\toprule
\textbf{Split} & \textbf{Sources} & \textbf{\#Dialogues} & \textbf{\#Tokens} & \textbf{\makecell{\#Candidate \\ Turns}} & \textbf{\makecell{\#Turns with \\ any PII}} & \textbf{\makecell{\%Turns with \\ NAME}} & \textbf{\makecell{\#Dialogues \\ without PII}} \\
\midrule
Train & Platform A & 800 & 950,886 & 92,735 & 1,295 & 83.7 & 314 \\
      & Platform B & 305 & 1,654,347 & 152,430 & 5,379 & 99.4 & 10 \\
Val   & Platform A & 100 & 134,006 & 12,980 & 242 & 75.6 & 43 \\
      & Platform B & 76 & 408,124 & 38,622 & 1,130 & 98.7 & 3 \\
Test  & Platform A & 100 & 111,313 & 11,606 & 165 & 81.2 & 39 \\
      & Platform B & 100 & 513,275 & 47,578 & 1,445 & 99.4 & 7 \\
Challenge Test & Platform A & 50 & 100,571 & 9,017 & 117 & 99.1 & 7 \\
               & Platform B & 80 & 490,421 & 39,005 & 2,743 & 100.0 & 0 \\
\midrule
\multicolumn{2}{l}{\textbf{Total}} & \textbf{1,611} & \textbf{4,362,943} & \textbf{403,973} & \textbf{12,516} & \textbf{97.1} & \textbf{423} \\
\bottomrule
\end{tabular}

\label{tab:dataset-summary}
\end{table}

\subsubsection{Canonical Evaluation Set}

The main evaluation set contains 200 held-out dialogues: 100 from Platform A and 100 from Platform B. We retain dialogues with zero gold PII spans (46 total: 39 from Platform A and 7 from Platform B) as negative examples, since a practical de-identification system must correctly decide when no redaction is necessary. These cases are important for measuring over-redaction and curricular preservation rather than recall alone.

The evaluation is strongly dominated by \textsc{Name} entities. On the canonical test set, 81.2\% of gold spans in Platform A and 99.4\% in Platform B are annotated as \textsc{Name}. This reflects the real operational distribution of educational dialogue, where student names, tutor names, and family references are the most common privacy-sensitive mentions.

\subsubsection{Challenge Set of Ambiguous Names}

A central failure mode in educational dialogues is ambiguity between real student identities and curricular content-based names often used in mathematical word problems. This distinction cannot be resolved reliably through lexical cues alone. To isolate this failure mode, we construct a targeted challenge set of 130 dialogues (50 from Platform A and 80 from Platform B) in which real student names and curricular-content-based names explicitly co-occur. The set was assembled by manually screening approximately 300 candidate dialogues for this co-occurrence pattern. These dialogues were annotated using the same guidelines and annotator process as the canonical test set, but they are intended as a stress test rather than a representative deployment sample. Challenge-set dialogues were excluded from all training splits, and no hyperparameter decisions were made using challenge-set results. Table~\ref{tab:dataset-summary} summarizes the dataset composition.

\subsection{Experimental Design}

\subsubsection{Baseline Conditions: LLM-Only Full-Dialogue Detectors}
\label{sec:fully-llm-baselines}

To establish baselines, we evaluated four LLM-only full-dialogue detectors, all using the same candidate-free span-extraction formulation. Each model received the entire tutoring dialogue as a single input example and directly produced a JSON-formatted \texttt{redact\_spans} output. The four baselines included: (1) Gemma E4B-it (4B), trained with LoRA using data specified in Table~\ref{tab:dataset-summary}; (2) Gemma-4-31B-it zero-shot prompting; (3) Gemma-4-31B-it with LoRA fine-tuning for one epoch using the same data specified in Table~\ref{tab:dataset-summary}; and (4) Gemini 3.1 Pro Preview \cite{google2026gemini3pro}, evaluated as a commercial API baseline in zero-shot mode through a secure institutional LiteLLM gateway under a formal data use agreement (DUA). The three Gemma variants ran on local hardware and therefore satisfied the privacy and governance requirements of local deployment. The Gemini comparison was included only as a controlled external reference point under DUA-authorized data egress, rather than as a deployable baseline.

\subsubsection{Experimental Conditions: Two-Stage Cascaded Framework}
\label{sec:experimental-conditions}

For RQ1, the base model was held constant --- Gemma E4B-it and Gemma-4-31B-it appear in both cascade and LLM-only conditions --- to isolate formulation from model scale. These formulations were evaluated on both the conical and challenge datasets (RQ2). For RQ3 (deployment footprint), we measured wall-clock training time, inference throughput, and resident memory on a single laptop. RoBERTa and 31B LoRA reviewers were trained with three random seeds (13, 42, 101); 4B LoRA and 31B LLM-only baselines used a single seed (42). The 4B reviewer was trained with a single seed due to compute budget; we treat it as a point estimate. For multi-seed configurations, all values in Tables~\ref{tab:main-results} and~\ref{tab:challenge-set} are means across the three seeds; per-seed standard deviations are in Appendix~\ref{sec:seed-variance}, and hyperparameter details are in Appendix~\ref{sec:hyperparams}.

\subsection{Evaluation}
\label{sec:evaluation}
To evaluate the detection performance, we report span-level precision, recall, and F1. A proposed span counts as a true positive if its token span overlap with any gold PII span; we use overlap matching rather than exact match to avoid penalizing benign boundary variation such as titles, possessives, and tokenization differences (e.g., when gold contains 'Mr. Smith' and a model identifies 'Smith'). This criterion is applied consistently across all models and experimental settings.

\section{Results}
\label{sec:results}

\subsection{RQ1: Performance of Cascade vs. LLM-Only Configurations}

Table~\ref{tab:main-results} shows that the cascaded proposer--reviewer formulation consistently outperformed LLM-only full-dialogue extraction when the base model family was held constant. This supports the central claim that in tutoring dialogue de-identification, problem formulation matters more than model scale alone.

\begin{table}[!ht]
\caption{Performance on the canonical evaluation test. ``Union'' = DeBERTa-v3-base + ModernBERT-base proposer candidates with direct RegEx rules (see Section~\ref{sec:stage1} for details).}
\label{tab:main-results}
\centering
\small
\setlength{\tabcolsep}{4pt}
\begin{tabular}{@{}lccccccc@{}}
\toprule
\multirow{2}{*}{\textbf{Configuration}}
& \multicolumn{3}{c}{\textbf{Platform A}}
& \multicolumn{3}{c}{\textbf{Platform B}}
& \multirow{2}{*}{\textbf{Macro F1}} \\
\cmidrule(lr){2-4} \cmidrule(lr){5-7}
& \textbf{Precision} & \textbf{Recall} & \textbf{F1}
& \textbf{Precision} & \textbf{Recall} & \textbf{F1}
& \\
\midrule
\emph{Proposer-Only} \\
\ \ DeBERTa & .867 & .946 & .904 & .959 & .977 & .968 & .936 \\
\ \ Union (DeBERTa \& ModernBERT)  & .714 & .970 & .823 & .878 & .981 & .927 & .875 \\
\emph{Proposer + Reviewer} \\
\ \ Union + RoBERTa & .841 & .960 & .896 & .953 & .977 & .965 & .931 \\
\ \ Union + Gemma E4B & .859 & .994 & .921 & .956 & .972 & .964 & .943 \\
\ \ \textbf{Union + Gemma 31B} & \textbf{.926} & \textbf{.978} & \textbf{.951} & \textbf{.952} & \textbf{.977} & \textbf{.965} & \textbf{.958} \\
\midrule
\emph{LLM-Only} \\
\ \ Gemma E4B (0-shot) & .880 & .400 & .550 & .950 & .120 & .213 & .381 \\
\ \ Gemma 31B (0-shot) & 1.000 & .564 & .721 & .993 & .093 & .170 & .446 \\
\ \ Gemma 31B LoRA & .913 & .891 & .902 & .952 & .474 & .632 & .767 \\
\ \ Gemini 3.1 Pro & .962 & .770 & .855 & .984 & .389 & .557 & .706 \\
\emph{Industry-Standard} \\
\ \ Presidio (default) & .153 & .824 & .258 & .299 & .906 & .450 & .354 \\
\ \ spaCy (en\_core\_web\_sm) & .031 & .770 & .060 & .051 & .911 & .097 & .078 \\
\bottomrule
\end{tabular}
\end{table}

The union proposer behaves as intended: it improves recall relative to DeBERTa alone while reducing precision, shifting precision recovery to the reviewer stage.

Adding a reviewer recovered most of this precision loss while preserving high recall. The best overall system was \emph{Union + Gemma 31B}, achieving the highest macro F1 of 0.958 (0.951 on Platform A, 0.965 on Platform B). The 4B reviewer also performed strongly (0.943 macro F1), showing that strong local performance does not require the largest model.

In contrast, LLM-only baselines showed severe recall collapse, especially on Platform B. The 4B LLM-only baseline reached only 0.381 macro F1, and even Gemma 31B LoRA reached 0.767 --- still 0.191 below the 31B cascade reviewer. Gemini 3.1 Pro improved to 0.706 but remained below all cascade systems while requiring external data egress. This collapse correlates with dialogue length: per-dialogue recall for the 4B LLM-only baseline drops from 0.54 on dialogues under 2K characters to 0.11 on dialogues over 20K characters (Appendix B).

Industry-standard local baselines (Presidio and spaCy) achieved high recall but extremely low precision, heavily over-redacting curricular content. Overall, a recall-first proposer followed by binary \redact{}/\keep{} review proved substantially more effective than direct full-dialogue extraction.

\subsection{RQ2: Performance on The Challenge Set of Ambiguous Names}
\label{sec:challenge-set}

Table~\ref{tab:challenge-set} evaluates the failure mode typical to educational context: ambiguous names that may refer either to real student identities (\redact) or to curricular-content-based names (\keep).

\begin{table*}[htbp]
\caption{Performance on the challenge set of ambiguous names.}
\label{tab:challenge-set}
\centering
\small
\setlength{\tabcolsep}{4pt}
\begin{tabular}{@{}lccccccc@{}}
\toprule
\multirow{2}{*}{\textbf{Configuration}}
& \multicolumn{3}{c}{\textbf{Platform A}}
& \multicolumn{3}{c}{\textbf{Platform B}}
& \multirow{2}{*}{\textbf{Macro F1}} \\
\cmidrule(lr){2-4} \cmidrule(lr){5-7}
& \textbf{Precision} & \textbf{Recall} & \textbf{F1}
& \textbf{Precision} & \textbf{Recall} & \textbf{F1}
& \\
\midrule
\emph{Proposer-Only} \\
\ \ DeBERTa-only
& .408 & .966 & .574
& .985 & .871 & .925
& .750 \\

\emph{Proposer + Reviewer} \\
\ \ Union + RoBERTa
& .425 & .912 & .580
& .975 & .654 & .784
& .682 \\

\ \ Union + Gemma E4B
& .410 & .974 & .577
& .872 & .991 & .928
& .753 \\

\ \ \textbf{Union + Gemma 31B}
& \textbf{.899} & \textbf{.892} & \textbf{.894}
& \textbf{.994} & \textbf{.945} & \textbf{.969}
& \textbf{.932} \\

\emph{LLM-Only}\\
\ \ Gemma E4B
& .457 & .319 & .376
& .960 & .157 & .270
& .323 \\

\ \ Gemini 3.1 Pro
& .644 & 1.000 & .784
& .988 & .939 & .963
& .873 \\
\emph{Industry-Standard} \\
\ \ Presidio (default) & .099 & .761 & .175 & .360 & .883 & .512 & .348 \\
\ \ spaCy (en\_core\_web\_sm) & .022 & .761 & .044 & .080 & .883 & .147 & .096 \\
\bottomrule
\end{tabular}
\end{table*}

The strongest system was again \emph{Union + Gemma 31B}, which achieved the best macro F1 of 0.932. On Platform A, it improved F1 from 0.574 (DeBERTa proposer only) to 0.894, mainly by recovering precision: proposer-only precision was only 0.408 due to heavy over-redaction, while the 31B reviewer raised it to 0.899 while maintaining strong recall (0.892). On Platform B, it also achieved the best result (0.969 F1), showing that contextual review is essential for resolving ambiguous names.

Smaller reviewers were much less robust. \emph{Union + RoBERTa} and \emph{Union + Gemma E4B} remained near 0.58 F1 on Platform A, showing little improvement over proposer-only detection. RoBERTa also reduced recall on Platform B (0.654), indicating that encoder reviewers often compounded proposer errors rather than correcting them.

The LLM-only 4B baseline performed worst overall (0.323 macro F1) because of severe recall collapse. The commercial API baseline (\emph{Gemini 3.1 Pro}) performed strongly (0.873 macro F1), but still underperformed the fully local 31B cascade, especially on Platform A (0.784 vs.\ 0.894 F1).

\begin{figure}[!ht]
\centering
\includegraphics[width=0.85\linewidth]{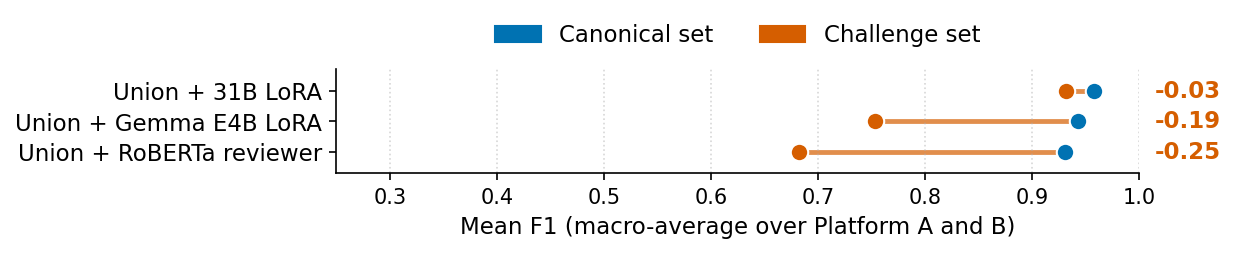}

\caption{Canonical-to-challenge F1 shift per configuration.}
\label{fig:challenge-degradation}
\end{figure}

Canonical-to-challenge degradation (Figure~\ref{fig:challenge-degradation}) further highlights the robustness of \emph{Union + Gemma 31B}: it drops by only .03 F1, compared with 0.19 for Gemma E4B LoRA and 0.25 for RoBERTa. It is the only configuration that maintains near-canonical performance, breaking the apparent near-equivalence among reviewers on the canonical test and establishing the 31B reviewer as the most robust reviewer by a substantial margin.

These results show that ambiguous-name resolution is not primarily a span-detection problem. Strong performance requires contextual \redact{}/\keep{} review rather than better candidate extraction alone.

\subsection{RQ3: Deployment Footprint}

To assess deployment feasibility, we measured wall-clock training time, inference throughput, and resident memory for each configuration on a single Apple M5 Max with 128\,GB unified memory. Table~\ref{tab:systems} summarizes these costs alongside macro F1 for reference.

\begin{table}[htbp]
\caption{Training and inference cost on Apple M5 Max (128\,GB, warm-start). Training time is wall-clock per seed. 31B cascade dial/s is reviewer-stage timing on the 200-dialogue canonical workload (proposer + rules add $<$5\,s, negligibly).}
\label{tab:systems}
\centering
\small
\setlength{\tabcolsep}{4pt}
\begin{tabular}{@{}lcccc@{}}
\toprule
\textbf{Configuration} & \textbf{Training Time (h)} & \textbf{\makecell{\ \ Throughput \\ (\#Dialogues/s)}} & \textbf{\makecell{Inference \\ RSS (GB)}} & \textbf{F1} \\
\midrule
\emph{Proposer-Only}\\
\ \ DeBERTa-only & $\sim$2.1 & 0.52 & 4.7 & .936 \\
\emph{Proposer + Reviewer}\\
\ \ Union + RoBERTa & $\sim$1.3 & 0.50 & 6.1 & .931 \\
\ \ Union + Gemma E4B & $\sim$2.8 & 0.36 & 9.5 & .943 \\
\ \ Union + Gemma 31B & $\sim$4.3 & 0.09 & $\sim$18 & .958 \\
\emph{LLM-Only}\\
\ \ Gemma E4B & 1.2 & 0.76 & 5.5 & .381 \\
\ \ Gemma 31B LoRA & 1.8 & 0.06 & $\sim$18 & .767 \\
\bottomrule
\end{tabular}
\end{table}

Table 4 shows that the best system remains feasible on a single Apple M5 Max: Union + Gemma 31B reaches 0.958 F1 with approximately 4.3 hours of training per seed and 18 GB inference memory. The 4B reviewer offers a strong efficiency tradeoff, retaining most of the accuracy at roughly half the memory and higher throughput. LLM-only models are less efficient in accuracy-per-resource terms: Gemma 31B LoRA uses similar memory to the 31B cascade but reaches only 0.767 F1. Overall, deployment cost is shaped less by model size alone than by whether model capacity is applied to localized Redact/Keep decisions.

\section{Conclusion}

We introduced a fully local cascade framework for de-identifying educational dialogue. The framework separates the task into high-recall candidate proposal followed by contextual Redact/Keep review. This formulation addresses a central challenge in educational data mining: preserving the analytic utility of large-scale learning data while satisfying privacy and governance constraints.

Across math tutoring transcripts from two large platforms, the cascade substantially outperformed same-family LLM-only extraction and a commercial API baseline. The strongest configuration, Union + Gemma 31B, reached 0.958 macro F1 on the canonical test set, compared with 0.767 for the same model used as a single-pass detector and 0.706 for the commercial API baseline. On a targeted challenge set for curricular–personal name ambiguity, the same configuration degraded by only 0.03 F1, while smaller reviewers degraded by 0.19 to 0.25 F1. These results suggest that, in this setting, problem formulation can matter more than model scale alone: strong de-identification requires not only finding candidate spans, but deciding whether those spans represent real privacy risk in context.

The deployment results further show that privacy-preserving de-identification need not require data egress to third-party APIs. The strongest configuration trained and ran on a single laptop-class machine, requiring approximately 4.3 hours of training per seed and 18 GB of inference memory. This makes the approach practical for institutions that need to process sensitive educational dialogue under local governance constraints.

Several limitations remain: First, the evaluation is limited to English-language math tutoring transcripts from two platforms; performance may differ in other subjects, languages, age groups, or instructional settings. Second, the cascade adds inference latency relative to a single-pass detector, which may matter for real-time applications even if it is acceptable for batch de-identification. Third, deployment-cost estimates are specific to Apple Silicon with unified memory and may vary on discrete-GPU infrastructure. Finally, the challenge set is intentionally targeted and relatively small, so its results should be interpreted as a stress test rather than a representative deployment estimate.

Future work should evaluate the framework across broader educational domains, expand challenge sets for additional ambiguity types, and study human-in-the-loop review policies for low-confidence cases. More broadly, this work shows that educational de-identification is not simply a generic NER problem; it is a contextual privacy-triage problem in which the goal is to protect learners while preserving the educational content needed for research and improvement.

\begin{acknowledgments}
This material is based upon initial work completed under National Science Foundation Grant No.~2321499, and support from the Gates Foundation and the Chan Zuckerberg Initiative. Any opinions, findings, and conclusions or recommendations expressed in this material are those of the authors and do not necessarily reflect the views of the funders.
\end{acknowledgments}

\section*{Declaration on Generative AI}
During the preparation of this work, the authors used ChatGPT (OpenAI) and Claude (Anthropic) in order to: Grammar and spelling check, Paraphrase and reword, and Improve writing style. After using these tools, the authors reviewed and edited the content as needed and take full responsibility for the publication's content.

\bibliography{references}

\clearpage
\appendix

\section{Methodology Notes}
\label{sec:methodology-notes}

Candidate spans are divided into two pools with different inference paths. (1) \textbf{Direct-rule pool} (18 rows on Platform A, 0 on Platform B): structured identifiers matched by deterministic rules (e.g., email, phone number, URL), which are emitted as \redact{} automatically, without reviewer inspection. (2) \textbf{Proposer pool} (212 rows on Platform A, 1,615 on Platform B): candidate spans generated by the union of DeBERTa + ModernBERT proposer, which are routed to the reviewer for binary \redact/\keep classification. Both pools contribute to span-level precision, recall, and F1 in Table~\ref{tab:main-results}. Direct-rule candidates are emitted directly as final output, bypassing the reviewer.

\section{Length Analysis}
\label{sec:length-analysis}

\begin{figure}[!ht]
\caption{Per-dialogue recall vs.\ length for the fully-LLM Gemma E4B baseline on 156 canonical dialogues with $\geq$1 gold span.}
\label{fig:recall-vs-length}
\centering
\includegraphics[width=0.75\linewidth]{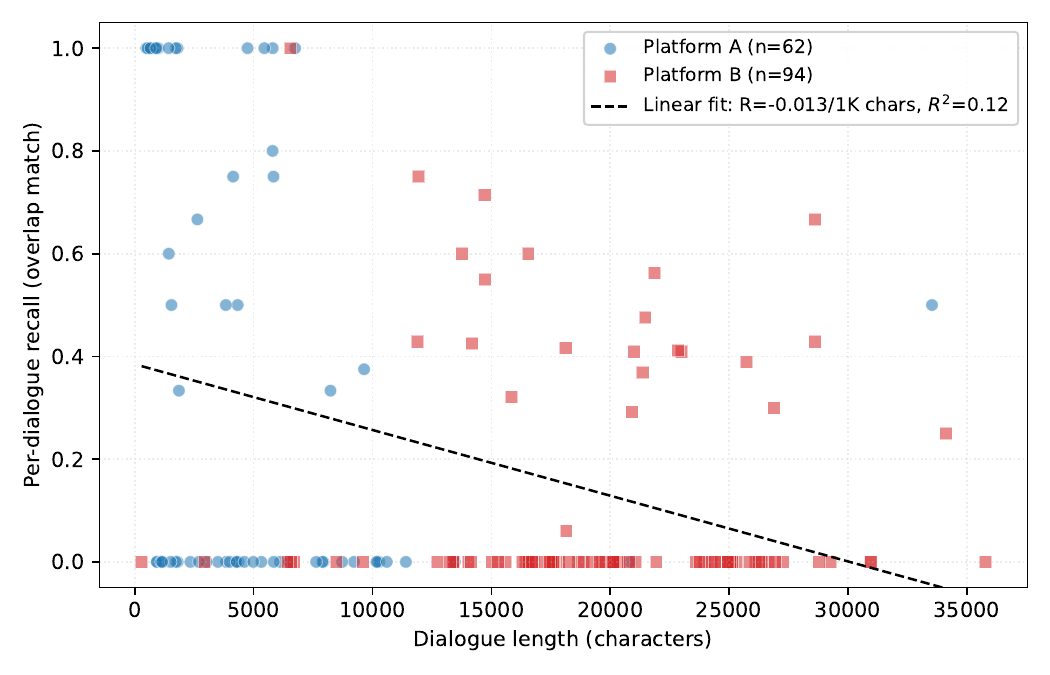}
\end{figure}

Figure~\ref{fig:recall-vs-length} shows that per-dialogue recall for the fully-LLM Gemma E4B baseline declines as dialogue length grows: a linear fit yields slope $-0.013$ per 1K characters ($p < 10^{-5}$, $R^2 = 0.12$), and bucketed mean recall drops from 0.54 ($<$2K chars) to 0.11 ($\geq$20K chars). We hypothesize that this recall collapse reflects an attention-budget limitation: the full-dialogue formulation must locate every PII span across thousands of tokens in a single generation pass, whereas the cascade reviewer applies full contextual attention to each candidate span individually. Alternative explanations (output-length saturation, span-extraction format constraints, and multi-target coordination) are not ruled out.

\FloatBarrier

\section{Reviewer Hyperparameters}
\label{sec:hyperparams}

All reviewers were trained on cascade-aligned data derived from the union proposer over the Platform A and Platform B training dialogues (Table~\ref{tab:dataset-summary}), with the 100-dialogue Platform A validation set used to monitor loss during training. 
Hyperparameters were selected without using either canonical test set or the challenge set; therefore, Platform A/B differences in Table~\ref{tab:main-results} reflect held-out evaluation rather than test-set tuning.
All reviewers use the AdamW optimizer; the 31B 3-seed training additionally requires an mlx-lm gradient-clipping patch.

\begin{table}[!ht]
\caption{Reviewer training hyperparameters.}
\label{tab:hyperparams}
\centering
\small
\begin{tabular}{@{}llll@{}}
\toprule
\textbf{Parameter} & \textbf{RoBERTa (Full FT)} & \textbf{Gemma E4B (LoRA)} & \textbf{Gemma 31B (LoRA)} \\
\midrule
Model & roberta-base (125M) & gemma-4-E4B-it (4B) & gemma-4-31B-it (31B) \\
Quantization & none (fp16) & 4-bit (q4) & 4-bit (q4) \\
LoRA config & n/a & r=8, $\alpha$=20, d=0, 16L & r=8, $\alpha$=20, d=0, 16L \\
Learning rate & 5e-5 & 1e-5 & 1e-5 \\
Epochs & 3 & 3 & 1 \\
Eff.\ batch & 8 (4$\times$2) & 8 (8$\times$1) & 8 (4$\times$2) \\
Max seq len & 384 & 512 & 512 \\
Warmup & 0.1 ratio & none & none \\
Loss & focal ($\gamma$=2.0) & CE (mask-prompt) & CE (mask-prompt) \\
Labels & REDACT/KEEP& REDACT/KEEP & REDACT/KEEP \\
Seeds & 13, 42, 101 & 42 & \textbf{13, 42, 101} \\
Training rows & 7,527 & 7,527 & 7,527 \\
\bottomrule
\end{tabular}
\end{table}
\FloatBarrier

\section{Per-Seed Variance for Multi-Seed Configurations}
\label{sec:seed-variance}

Tables~\ref{tab:main-results} and~\ref{tab:challenge-set} report 3-seed means for the \emph{Union + RoBERTa} and \emph{Union + Gemma 31B} configurations. Table~\ref{tab:seed-variance} below reports the corresponding per-seed standard deviations across seeds 13, 42, and 101. Span-level F1 is stable across seeds ($<$0.01 std on canonical surfaces); precision and recall on the Platform A challenge set show the largest variance, reflecting prompt-sensitivity × LoRA-adapter interaction on the small (50-dialogue) word-problem ambiguity probe.

\begin{table}[!ht]
\caption{Per-seed standard deviations for the multi-seed configurations. Values are reported as $\textit{mean} \pm \textit{std}$ across seeds 13, 42, and 101.}
\label{tab:seed-variance}
\centering
\small
\setlength{\tabcolsep}{4pt}
\begin{tabular}{@{}llcccc@{}}
\toprule
\textbf{Configuration} & \textbf{Set} & \textbf{Platform} & \textbf{Precision} & \textbf{Recall} & \textbf{F1} \\
\midrule
\multirow{4}{*}{Union + RoBERTa}
& Canonical & A & .841{\scriptsize$\pm$.014} & .960{\scriptsize$\pm$.004} & .896{\scriptsize$\pm$.009} \\
& Canonical & B & .953{\scriptsize$\pm$.002} & .977{\scriptsize$\pm$.001} & .965{\scriptsize$\pm$.001} \\
& Challenge & A & .425{\scriptsize$\pm$.014} & .912{\scriptsize$\pm$.030} & .580{\scriptsize$\pm$.009} \\
& Challenge & B & .975{\scriptsize$\pm$.003} & .654{\scriptsize$\pm$.011} & .784{\scriptsize$\pm$.008} \\
\midrule
\multirow{4}{*}{Union + Gemma 31B}
& Canonical & A & .926{\scriptsize$\pm$.021} & .978{\scriptsize$\pm$.019} & .951{\scriptsize$\pm$.009} \\
& Canonical & B & .952{\scriptsize$\pm$.008} & .977{\scriptsize$\pm$.001} & .965{\scriptsize$\pm$.004} \\
& Challenge & A & .899{\scriptsize$\pm$.036} & .892{\scriptsize$\pm$.035} & .894{\scriptsize$\pm$.010} \\
& Challenge & B & .994{\scriptsize$\pm$.001} & .945{\scriptsize$\pm$.004} & .969{\scriptsize$\pm$.002} \\
\bottomrule
\end{tabular}
\end{table}

\section{Reviewer Prompt}
\label{sec:reviewer-prompt}

Both the 4B and 31B LoRA reviewers share a curricular-content-aware system prompt.

\paragraph{Design rationale.} A prompt that treats every candidate as suspicious by default over-redacts names used inside math word problems (e.g., ``Jordan has 15 apples''), dragging down F1 on the A challenge set: 3-seed mean F1 drops from 0.680 (without a word-problem carve-out) to 0.894 (with the carve-out), a +0.21 absolute improvement. The final prompt adds two components to the suspicious-by-default baseline:
\begin{enumerate}
\itemsep0em
\item A word-problem note instructing \keep when a candidate name appears in a mathematical scenario.
\item Pedagogical direct-address counter-examples (``Hi, I'm Morgan'', ``Thanks, Taylor!'') using names chosen specifically to not overlap with the test-set failure cases, to avoid overfitting the prompt to specific evaluation spans.
\end{enumerate}

\paragraph{Prompt text (verbatim).}

\begin{verbatim}
You are a PII verification analyst for educational
de-identification.

You will receive one candidate span proposed by an
earlier detector. Treat the candidate as suspicious
by default.

Decide whether the transcript should REDACT or KEEP
the span.

Label semantics:
- REDACT: the span is real PII or should be removed
  for privacy.
- KEEP: the span is not PII and should remain in the
  transcript.

Only output KEEP if you are very confident the
candidate is non-PII educational or public content,
such as:
- a mathematical expression, variable, equation, or
  formula
- a historical figure, scientist, mathematician, or
  other public figure
- a fictional character, brand name, or product name
- a course name, textbook reference, section number,
  or assignment label
- a date used as a schedule, due date, semester, or
  other non-personal time reference
- a score, grade, percentage, or number used in an
  educational or math context
- a country, state, or region mentioned as general
  context rather than a personal address
- a theorem, method, law, or named concept whose
  person-name is part of the public concept

Always output REDACT for:
- real participants, tutors, students, teachers, or
  family members
- greetings, direct address, self-introductions, or
  other real interpersonal references
- schools, local places, usernames, links, contact
  details, or other direct identifiers

When in doubt, output REDACT.
Output exactly one label: REDACT or KEEP. Do not
output JSON or explanations.

Note: In math tutoring, names inside word problems or
story problems are fictional characters, not real
people. If the candidate span is a name used in a
mathematical scenario (e.g., "Jordan has 15 apples",
"Riley measured the length"), output KEEP.

However, names in greetings, self-introductions
(e.g., "Hi, I'm Morgan"), direct address ("Thanks,
Taylor!"), or any non-math context are real
participants -- output REDACT.
\end{verbatim}

\section{Error Analysis: Shortcut-Driven Reviewer Behavior}
\label{sec:mechanism-analysis}

A targeted audit of the RoBERTa reviewer shows it succeeds as a narrow shortcut-driven pruner, not a contextual understander. Its correct \keep decisions concentrate on three categories: story-problem character names (e.g., \emph{Tara}, \emph{Lucy}), obvious lexical false positives on curricular terms (\emph{Algebra}, \emph{Geometry}), and well-known public concepts (\emph{Venn}, \emph{Manchester City}).

A comparison of DeBERTa-v3-base and ModernBERT-base \cite{warner2024modernbert} at the proposer stage confirms the same mechanism: ModernBERT's lower precision (0.90 vs.\ DeBERTa's 0.98 on Platform B) is overwhelmingly composed of false positives on the same shortcut categories (98/132 are capitalized single-token story-problem names).

The shortcut-driven calibration is \textbf{double-edged}: on the canonical surface it gives DeBERTa cleaner false-positive suppression; on word-problem-heavy content it causes over-suppression of real PII (the Isaac case: 10 of 56 instances of the same name missed on one dialogue).

\section{Overlap-Matching Coverage Audit}
\label{sec:overlap-audit}

To test whether overlap matching masks residual leakage, we re-examined every gold span counted as a true positive and checked whether the union of predicted \redact spans in that turn \emph{fully} covers the gold span's tokens; any uncovered token is identifier text that would survive redaction. We applied the same augmenting-path span matching used for the main results.

For the final cascade (\emph{Union + Gemma 31B}), all 12{,}806 matched
redactions across three seeds and all four evaluation surfaces fully cover their gold spans --- zero partial covers. Overlap-based and full-coverage scoring are therefore identical for our system, and the reported recall does not benefit from boundary leniency. The off-the-shelf baselines show a small fraction of partial covers (Presidio 1.7\%, spaCy 2.0\% of matched spans on the canonical surfaces). These are almost entirely benign title prefixes (``Mr.''/``Ms.'' left visible while the surname is redacted) and URL/path fragments; the sole genuinely under-covered identifier is one phone number truncated by spaCy (``555'' matched, ``987-6543'' left visible). The leniency thus modestly \emph{inflates the baselines'} recall.

\end{document}